\documentclass{article}
\usepackage{spconf,graphicx}
\usepackage{times}
\usepackage{amsmath}
\usepackage{latexsym}
\usepackage{mathtools}
\usepackage{mathabx}
\usepackage{subcaption}
\usepackage{tikz}
\usepackage[inline]{enumitem}
\usepackage{paralist}       \usepackage{graphicx}
\usepackage{hhline}

\usepackage{blindtext}
\usepackage{hyperref}

\usepackage{tabto}
\usepackage{xspace}
\usepackage{multirow}
\newcommand{\method}{{UniPunc}\xspace}
\makeatletter
\DeclareRobustCommand\onedot{\futurelet\@let@token\@onedot}
\def\@onedot{\ifx\@let@token.\else.\null\fi\xspace}
\def\eg{\emph{e.g}\onedot} 
\def\ie{\emph{i.e}\onedot} 
 
 \def\vs{\emph{vs}\onedot}

\newcommand{\veca}{\ensuremath{\mathbf{a}}\xspace}
\newcommand{\vecx}{\ensuremath{\mathbf{x}}\xspace}
\newcommand{\vecy}{\ensuremath{\mathbf{y}}\xspace}

\newcommand{\vecHa}{\ensuremath{\mathbf{H}_a}\xspace}
\newcommand{\vecHl}{\ensuremath{\mathbf{H}_l}\xspace}
\newcommand{\vecHi}{\ensuremath{\mathbf{H}_i}\xspace}

\newcommand{\vecH}{\ensuremath{\mathbf{H}}\xspace}

\newcommand{\vecSl}{\ensuremath{\mathbf{S}_l}\xspace}
\newcommand{\vecSa}{\ensuremath{\mathbf{S}_a}\xspace}

\newcommand{\att}{\ensuremath{\mathrm{Att}}\xspace}
\newcommand{\softmax}{\ensuremath{\mathrm{softmax}}\xspace}

\title{Unified Multimodal Punctuation Restoration Framework for Mixed-Modality Corpus}
\name{Yaoming Zhu\thanks{\ \ Yaoming and Liwei have equal contribution},~Liwei Wu,~Shanbo Cheng,~Mingxuan Wang}
\address{ByteDance AI Lab \\
\{zhuyaoming, wuliwei.000, chengshanbo,wangmingxuan.89\}@bytedance.com
}

\begin{document}
%
\maketitle 

\begin{abstract}
The punctuation restoration task aims to correctly punctuate the output transcriptions of automatic speech recognition systems. Previous punctuation models, either using text only or demanding the corresponding audio, tend to be constrained by real scenes, where unpunctuated sentences are a mixture of those with and without audio. This paper proposes a unified multimodal punctuation restoration framework, named \method, to punctuate the mixed sentences with a single model. \method jointly represents audio and non-audio samples in a shared latent space, based on which the model learns a hybrid representation and punctuates both kinds of samples. We validate the effectiveness of the \method on real-world datasets, which outperforms various strong baselines~(\eg{~BERT, MuSe}) by at least 0.8 overall F1 scores, making a new state-of-the-art. Extensive experiments show that \method's design is a pervasive solution: by grafting onto previous models, \method enables them to punctuate on the mixed corpus. Our code is available at \url{github.com/Yaoming95/UniPunc}
\end{abstract}
\section{Introduction}

Automatic speech recognition~(ASR) has wide application and serves multiple tasks as an upstream component, like voice assistants and speech translations.
Typically, ASR output unsegmented transcripts without punctuations, which may lead to misunderstanding for people \cite{DBLP:conf/interspeech/TundikSGB18}, and affect performance of downstream modules, such as machine translation~\cite{vandeghinste2018comparison} and  information extraction \cite{DBLP:conf/interspeech/MakhoulBBNRSSX05}.

To address the issue, researchers proposed the automatic punctuation restoration task and designed a series of models. Conventional punctuation restoration models had achieved considerable progress~\cite{gravano2009restoring,tilk2015lstm,DBLP:conf/iwslt/CourtlandFM20,makhija2019transfer,DBLP:conf/icpr/WangCYX18}, but they were solely based on the lexical information, which gives rise to some problems. One sentence may have varied punctuation, contributing to different meanings, respectively. For example, ``I don't want anymore kids'' means far away from ``I don't want anymore, kids'', suggesting the importance of a comma. Additionally, unaware of the speaker's tone, the model might find it difficult to determine whether a sentence should end in a full stop or a question mark.

Acoustic signals can help mitigate the ambiguity of punctuation models, considering rich information in speech audio such as pauses and intonation. Therefore, \cite{klejch2017sequence,DBLP:conf/icassp/YiT19,DBLP:conf/interspeech/SunkaraRBBK20} proposed several multimodal punctuation restoration models. Generally, these multimodal models extracted acoustic features from speech audio and fused acoustic and lexical features by addition/concatenating. Their experiments validated that the speech audio of the text benefits punctuation.

Despite their effectiveness, previous multimodal systems faced with \textit{modality missing} in real applications. Firstly, storage constraints or privacy policies may deny access to the corresponding audio, where previous multimodal systems failed to punctuate such audio-free sentences;
Secondly, human-labeled punctuation audio is expensive and hard to obtain, which leads to sparser training sets for those multimodal models.
Meanwhile, the audio-free unpunctuated text is readily available.
Hence, an ideal punctuation model should utilize and punctuate both audio text and audio-free text.

To handle the modality missing problem, this paper proposes the \method, a unified multimodal punctuation framework enabling punctuation on both audio and audio-free text. Specifically, for audio text, \method first converts the two modalities into embedding by pretrained lexical model and acoustic model, while for the audio-free sentences, \method introduces virtual embedding to simulate its audio. \method then applies an attention-based module, named coordinated bootstrapper, to construct a cross-modal hybrid representation. Based on the hybrid representation, the model learns and predicts multimodal punctuation. The \method combines the strengths of the previous unimodal and multimodal models: it can utilize a large amount of audio-free corpora as the training set and take advantage of acoustic input to reduce ambiguity in punctuation. Thus, the model is applicable to the punctuation of both audio texts and non-audio texts.

We highlight our main contributions as follows:
\begin{inparaenum}[\it a)]
\item We propose \method, a new framework that serves punctuation restoration task and utilizes both audio-free text corpus and audio speech transcripts. 
\item \method achieves state-of-the-art performance on multimodal punctuation and outperforms various strong baselines by at least 0.8\% overall F1 scores on real-world datasets.
\item We discuss pervasiveness \method: the introduced framework can be grafted into related work~\cite{tilk2015lstm,silva2021multimodal}, empowering them to process both audio and non-audio text while improving performance. 
\end{inparaenum}

\section{Problem Formulation}

\label{sec:challenge}
Punctuation restoration is often modeled as a sequence labeling task~\cite{DBLP:conf/interspeech/ZelaskoSMSCD18}.
Generally, the corpus of multimodal punctuation is a set of sentence-audio-punctuation triples and denoted as $S=\{\vecx, \veca, \vecy\}$. Here $\vecx$ is an unpunctuated sentence of length $T$, and $\veca$ is the corresponding speech audio. The model should predict the sequence of punctuation $\vecy$ given $\vecx$ and $\veca$. The length of punctuation $\vecy$ is identical to the unpunctuated sentence $\vecx$ due to the nature of the sequence labeling task, \ie, $|\vecx| = |\vecy|$.
This study focuses on four types of punctuation, namely comma(,), full stop(.), question mark(?), and no punctuation.

As we have mentioned, our training set is a mix of audio text and audio-free text: on the one hand, we want to leverage large amounts of plain text in the absence of audio; on the other hand, we wish to make full use of acoustic features in the presence of audio. So is the case for the test set since we do not always get audio of the ASR transcription to be punctuated. 
Hence, the main challenge lies in modality missing for both training samples and evaluation data: \ie, $\exists~\veca= \O$.
\begin{figure}[!t]
\centering
\includegraphics[width=.99\linewidth]{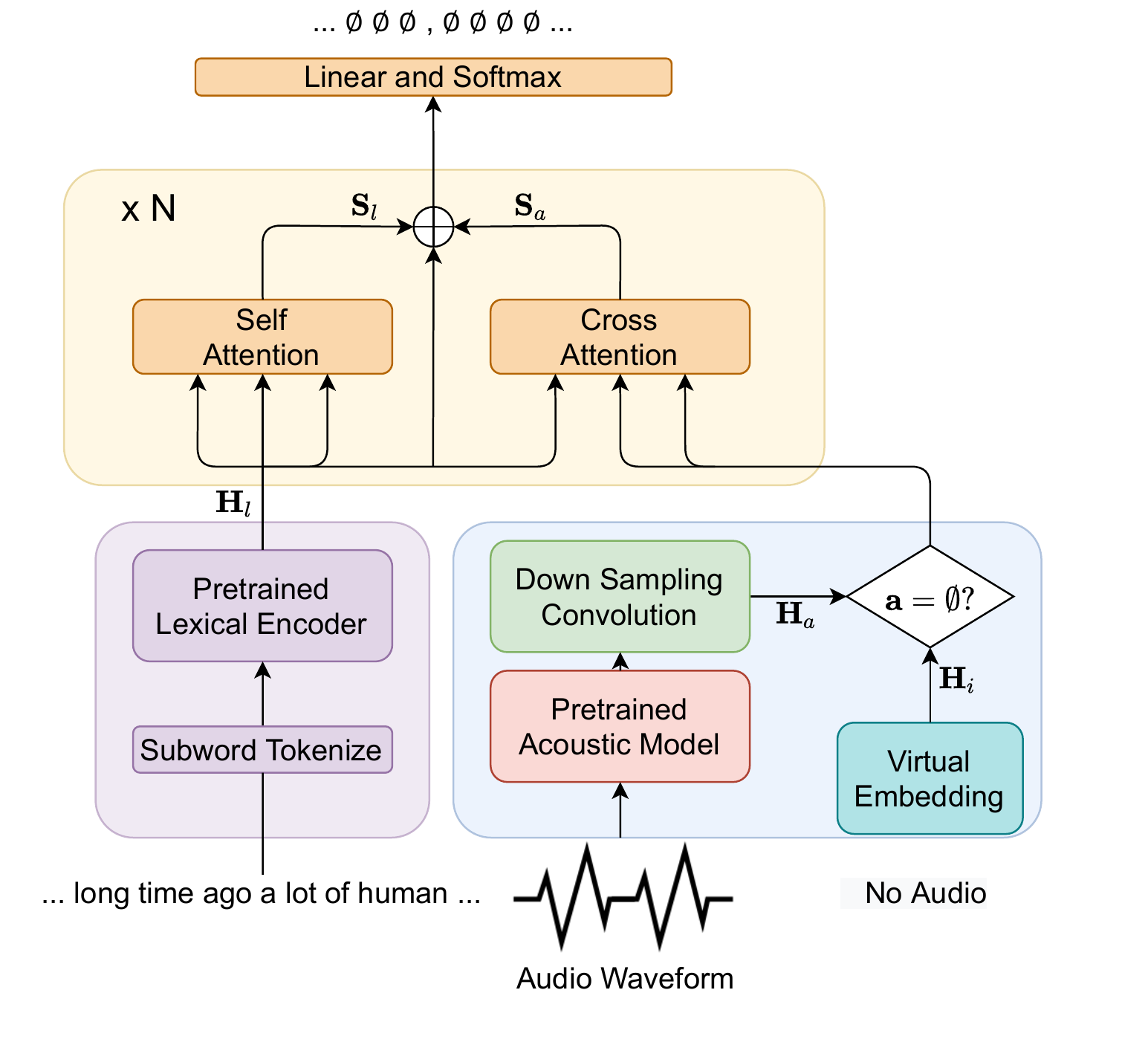}
\caption{Overall layout of \method. Best viewed in color. Lexical encoder, acoustic assistant, and coordinate bootstrapper are blocked in purple, blue, and yellow respectively.}
\label{fig:layout}
\end{figure}

\section{Method}
This section will introduce the proposed \method in detail. 
\method consists of three main components: lexicon encoder, acoustic assistant, and coordinate bootstrapper. The three components learn lexical features, possible acoustic properties, and cross-modal hybrid features, respectively. Fig.~\ref{fig:layout} shows the overall layout.


\subsection{Lexical Encoder}
The lexical encoder tokenizes unpunctuated sentence into subword sequence and converts it into the contextual embedding sequence $\vecHl=(h^l_1, ..., h^l_n)$. 
We leverage pretrained NLP model, e.g, BERT~\cite{DBLP:conf/naacl/DevlinCLT19} as our backbone model. We treat the pretrained model as a sub-module of \method and finetune the model on our task-specific data.


\subsection{Acoustic Assistant}
The acoustic assistant consists of an acoustic feature extractor, a down-sampling network, and a virtual embedding. It converts possible speech audio $\veca$ into acoustic embeddings $\vecHa$ or virtual embedding $\vecHi$. 

For the audio-annotated transcriptions, \method converts the audio signal into acoustic features via a pretrained acoustic feature extractor.
Generally, the acoustic feature extractor is first pretrained on unlabelled audio datasets through self-supervised training and is then used as a submodule of the whole model and finetuned on the downstream punctuation task.
Then \method applies a down-sampling network further to shorten the extracted acoustic features' length and get acoustic embedding $\vecHa=(h^a_1, ..., h^a_m)$. The purpose is to make the length of acoustic embeddings close to the sentence embedding so that the model can better align cross-modal information. We choose a multilayer convolutional network as the core component of the down-sampling network.

For the audio-free sentences, we propose to use virtual embedding $\vecHi$ simulating the possible missing acoustic feature, \ie, $\vecHa=\vecHi~\mathbf{if}~\veca=\O$.
We set virtual embedding as a fixed-length array of learnable parameters and expect it to learn the representations of absent audio. 
In the training process, the model learns the shared latent space of acoustic and virtual embedding, which supplies acoustic information for the subsequent coordinate bootstrapper.

\subsection{Coordinate Bootstrapper}

We propose the coordinate bootstrapper for jointly training audio-free and audio text to overcome the missing modality problem. The coordinate bootstrapper jointly exploits the acoustic and lexical features and applies an attention-based operation to learn a hybrid representation across two modalities.

Specifically, \method first conducts self-attention on the contextual embedding $\vecHl$ to capture the long-range dependencies $\vecSl$ within the unpunctuated sentence and apply cross-attention between the contextual embedding $\vecHl$ and acoustic embedding $\vecHa$ to formulate a cross-modal representation $\vecSa$:
\begin{equation}
   \vecSl =\att(\vecHl, \vecHl, \vecHl) 
\end{equation}
\vspace{-20pt}
\begin{equation}
   \vecSa =\att(\vecHa, \vecHa, \vecHl) 
\end{equation}
where $\att(q, k, v) = \softmax(\frac{q k^\top }{\sqrt{d_k}}) v$ is the attention operation proposed by \cite{vaswani2017attention}, and $d_k$ is the dimension size of the model. Note that for modality missing samples, we substitute the acoustic embedding $\vecHa$ with virtual embedding $\vecHi$, in which case  $\vecSa=\att(\vecHi, \vecHi, \vecHl)~\mathbf{if}~\veca=\O $.  

Then the \method acquires hybrid representation $\vecH_h$ by adding the attended representation along with a residual connection~\cite{he2016deep}. 
\begin{equation}
   \vecH_h = \vecSl + \vecSa + \vecHl
\end{equation}
Like other attention blocks, the coordinate bootstrapper can be stacked to multiple layers to increase the model capacity further.

Finally, the output classifier layer consists of a linear projection and a softmax activation function. \method inputs $\vecH$ to the classifier layer to predict the punctuation sequence $\hat{\vecy}$. 

In this way, we enable the representations of audio samples and no audio samples to share the same embedding space. Thus, the model can receive mixed data in the same training batch, and the trained model is able to punctuate both audio text and audio-free text.


It should be emphasized that \method makes a pervasive framework in solving modality missing in multimodal punctuation restoration tasks. For the previously proposed punctuation model, a proper adaptation of the proposed coordinate bootstrapper and acoustic assistant can empower them to address modality missing samples.
Section~\ref{sec:res} discuss the the pervasiveness of coordinate bootstrapper upon previous models~\cite{tilk2015lstm,silva2021multimodal} via experiment.

\section{Experiment}

\label{sec:exp}

\subsection{Datasets}
We conduct our experiment mainly on two real-world corpus: MuST-C~\cite{DBLP:journals/csl/CattoniGBNT21}\footnote{\url{https://ict.fbk.eu/must-c/}} and Multilingual TEDx(mTEDx)~\cite{DBLP:journals/corr/abs-2102-01757}\footnote{\url{http://www.openslr.org/100/}}, whose audio are both originated from TED talks.
We constructed three sets of data based on  two corpora:
\begin{inparaenum}[\it 1)]
\item English-Audio: This set contains the English audio and sentences in MuST-C. Each sample is with audio.
\item English-Mixed: This set contains all English audio sentences and audio-free sentences from two corpora. Note that English-Audio is a subset of English-Mixed.

\end{inparaenum}
\begin{table}[!h]

\caption{Statistical information of the training/test data on the two sets, where the first two rows are of the training set. The \textit{ sent. len} and \textit{ audio len.} denotes the \textit{average} length of sentence and audio counted by words and seconds, respectively.}
\label{Tab:Stat}
\resizebox{1\columnwidth}{!}{
\begin{tabular}{l|cccc}
\hline
              & \# of sent. & \# of audio &  sent. len. &  audio len. \\ \hline
English-Audio~(Train) & 99381       & 99381       & 42.5            & 14.7            \\
English-Mixed~(Train) & 142441      & 99381       & 44.1            & 14.7            \\ \hline
English-Audio~(Test) & 490         & 120940      & 54.7            & 15.7                \\
English-Mixed~(Test) & 2298        & 120940      & 52.0            & 15.7            \\ \hline
\end{tabular}
}

\end{table}

\begin{table*}[h]
\centering

\caption{Results for three punctuations on two test set.}
\label{Tab:results-on-english}
\resizebox{1.85\columnwidth}{!}
{
\begin{tabular}{c|l|cccccccccccc}
\hline
\multicolumn{1}{l|}{Test Set}                                                          & Model                        & \multicolumn{3}{c|}{Comma} & \multicolumn{3}{c|}{Full Stop} & \multicolumn{3}{c|}{Question Mark} & \multicolumn{3}{c}{Overall} \\ \hline
\multicolumn{1}{l|}{}                                                                  &                              & P       & R       & F1     & P        & R        & F1       & P          & R         & F1        & P       & R       & F1      \\ \hline
\multirow{7}{*}{English-Audio}                                                         & LSTM-T~\cite{DBLP:conf/interspeech/ZelaskoSMSCD18}                       & 68.8    & 50.7    & 56.9   & 75.8     & 74.9     & 74.3     & 29.8       & 26.5      & 27.2      & 72.0    & 60.0    & 65.1    \\
                                                                                       & Self-Att-Word-Speech~\cite{DBLP:conf/icassp/YiT19}         & 71.1    & 58.1    & 62.7   & 81.1     & 76.4     & 77.5     & 31.9       & 30.1      & 31.0      & 75.5    & 65.3    & 69.7    \\
                                                                                       & BERT~\cite{makhija2019transfer}                         & 79.1    & 66.9    & 72.1   & 84.5     & 81.7     & 82.8     & 78.4       & 75.7      & 75.5      & 81.5    & 73.5    & 77.1    \\
                                                                                       & MuSe~\cite{DBLP:conf/interspeech/SunkaraRBBK20}                         & 78.5    & 68.7    & 73.2   & 83.0     & 84.5     & 83.6     & 81.4       & 79.7      & 79.4      & 80.6    & 75.4    & 77.9    \\
                                                                                       & MuSe~\cite{DBLP:conf/interspeech/SunkaraRBBK20}+TTS~\cite{soboleva2021replacing}                     & 75.3    & 78.9    & 77.1   & 80.0     & 77.6     & 78.8     & 78.0       & 80.9      & 79.5      & 75.6    & 79.9    & 77.8    \\ \cline{2-14}
                                                                                       & \method-Audio~(Ours) & 78.9    & 70.2    & 74.2   & 81.7     & 86.0     & 83.7     & 84.3       & 80.1      & 80.8      & 80.3    & 76.8    & 78.5    \\
                                                                                       & \method-Mix~(Ours)   & 69.9    & 79.7    & 74.5   & 85.1     & 82.6     & 83.8     & 79.0       & 81.9      & 80.4      & 76.3    & 81.0    & 78.6    \\ \hhline{=|=|============}
\multirow{5}{*}{\begin{tabular}[c]{@{}c@{}}English-Mixed\end{tabular}} & BiLSTM~\cite{tilk2015lstm}                       & 61.0    & 45.0    & 51.6   & 59.5     & 54.7     & 56.8     & 58.8       & 47.7      & 50.1      & 60.5    & 48.3    & 53.7    \\
                                                                                       & Att-GRU~\cite{kim2019deep}                    & 61.1    & 47.5    & 52.0   & 75.7     & 66.9     & 69.6     & 27.3       & 26.5      & 26.0      & 67.1    & 55.5    & 60.3    \\
                                                                                       & SAPR~\cite{DBLP:conf/icpr/WangCYX18}                & 72.1    & 58.6    & 64.4   & 78.3     & 75.9     & 76.9     & 76.5       & 56.9      & 63.7      & 74.8    & 65.3    & 69.7    \\
                                                                                       & BERT~\cite{makhija2019transfer}                         & 74.2    & 68.0    & 70.8   & 82.6     & 81.2     & 81.8     & 78.7       & 79.5      & 77.4      & 78.0    & 73.6    & 75.7    \\ \cline{2-14}
                                                                                       & \method-Mix~(Ours)   & 73.2    & 71.2    & 72.1   & 82.5     & 82.2     & 82.2     & 79.0       & 76.0      & 76.1      & 77.3    & 75.8    & 76.5    \\ \hline
\end{tabular}
}
\end{table*}

We re-partition the data for each sample corresponding to an audio duration roughly range from 10 to 30 seconds. Tab.~\ref{Tab:Stat} shows statistical information of the text and audio about the training and test sets in detail.

\subsection{Configurations and Baselines}

We set most modules at a dimension of 768. 
We choose BERT~\cite{DBLP:conf/naacl/DevlinCLT19} and its subword tokenizer as pretrained lexical encoder and Wav2Vec 2.0~\cite{DBLP:conf/nips/BaevskiZMA20}\footnote{The pretrained acoustic model used in the MuSe was Wav2Vec, which we also replace it with a newer and better version of Wav2Vec 2.0 in our implement for a fair comparison with \method.} as pretrained acoustic model. For two English datasets, we use BERT-base uncased and Wav2Vec 2.0-base no tuned version.
We use a two-layer Convolution network of stride 5, kernel size of 15 as the down-sampling network. The layer number of coordinate bootstrapper is 2. The sequence length of virtual embedding is 5. We use learning rate of 0.00001 with Adam~\cite{DBLP:journals/corr/KingmaB14}, dropout rate of 0.1, and Noam learning rate scheduler of warm-up step 8000.

We compare the performance of \method with various baselines and SOTA systems, including:
LSTM-T~\cite{DBLP:conf/interspeech/ZelaskoSMSCD18}, Att-GRU~\cite{kim2019deep}, BERT~\cite{makhija2019transfer}, SAPR~\cite{DBLP:conf/icpr/WangCYX18}, Self-Att-Word-Speech\cite{DBLP:conf/icassp/YiT19} and MuSe~\cite{DBLP:conf/interspeech/SunkaraRBBK20}. We also compare to TTS punctuation data augmentation~\cite{soboleva2021replacing}.

All unimodal baselines are trained on English-Mixed corpus, while multimodal ones are trained on English-Audio since they cannot handle audio-free samples.
For \method, we first trained it jointly on English-Mix, denoted as \method-Mixed. To facilitate comparison with other multimodal baselines, we trained a \method variant solely on the English-Audio, denoted as \method-Audio.
All models and baselines are evaluated in terms of precision(P), recall(R), and F1 score(F1) on three punctuations. 

We implement \method by Fairseq~\cite{ott2019fairseq}. All experimental data are publicly available online. We will release our code and data split after paper acceptance for reproducibility. 


\section{Results}
\label{sec:res}

\subsection{Main Results}

We report the performance of our model and other baselines on English datasets in Tab.~\ref{Tab:results-on-english}.

\noindent \textbf{Compare with other multimodal models:} We compare \method with other baselines on the English-Audio test set and get the following main conclusions: 1.~\method-mixed's overall F1 score exceeds all baselines by at least 0.7 points, indicating the effectiveness of our framework. 2.~\method-Audio also outperforms all multimodal baselines, suggesting that even without introducing extra audio-free corpus, \method makes a strong model on multimodal sequence labeling; 3.~The overall F1 score of \method-Mix is slightly higher than that of \method-Audio, indicating that training with audio-free sentences indeed improves the performance, especially in punctuating commas and full stops. 4.~Using audio-free text mainly improves the recall of punctuation (compare Muse~\vs.~Muse+TTS, UniPunc-Audio~\vs.~UniPunc-Mixed)

\noindent \textbf{Compare with other unimodal models:} We also compare \method-mixed and other unimodal punctuation models on the English-Mixed test set; we also list BERT's performance on English-Audio as it is the strongest unimodal baseline. Our conclusions are twofold: 1.~\method outperforms all baselines with at least 0.8 overall F1 scores on English-Mixed, which proves that its learned hybrid representation is very effective; 2.~\method far outperforms BERT on English-Audio 1.5 overall F1 score, which shows that \method effectively represents the acoustic features in speech, which is especially obvious for punctuating question marks.

\begin{table}[]
\footnotesize
\caption{F1 score for pervasive experiment on English-Mixed test set. CD denotes content dropout. }
\label{Tab:results-on-pervasice-experiment}
\resizebox{0.9\columnwidth}{!}
{
\begin{tabular}{l|cccc}
\hline
             & Comma & Full Stop & Question Mark & Overall \\ \hline
BiLSTM~\cite{tilk2015lstm} & 51.6  & 56.8      & 50.1          & 53.7    \\
BiLSTM-\method  & 55.7  & 58.4      & 46.5          & 56.6    \\ 
CD~\cite{silva2021multimodal}-BERT   & 71.4  & 81.9      & 69.7          & 76.1    \\
CD-\method    & 71.5  & 82.1      & 76.7          & 76.9     \\ \hline
\end{tabular}
}
\end{table}

We also conduct a case study and evaluate \method's performance on multilingual data from mTEDx. We find \method punctuates closer to  humans, and can better tell the pauses of commas and fulls stops as well as the tone of questions. In addition, we also find that \method has better punctuation performance than other baselines on multilingual punctuation, which suggests that \method has better robustness and generalization. 
The examples of case study is available at our code base. 

\subsection{Pervasiveness of \method}



This subsection explores the pervasiveness of \method framework for punctuation restoration on the mixed datasets by grafting onto two previous approaches, namely BiLSTM~\cite{tilk2015lstm} and content dropout~\cite{silva2021multimodal}. Specifically, for the BiLSTM model, we introduce an acoustic assistant to extract potential acoustic features and coordinate bootstrapper to learn hybrid representation. We also graft \method with content dropout and compare it with the lexical content dropout BERT.
Tab.~\ref{Tab:results-on-pervasice-experiment} shows the overall results.

The experiments show that our framework is pervasive for solving the modality missing in punctuation, and enables the previous unimodal model to handle multimodal corpus. By grafting \method onto the BiLSTM, our module greatly improves the performance of the original model by 2.9 F1 scores.
In particular, when \method and content dropout are used jointly, the model achieves an F1 score of 76.9, further improving the overall performance.

\section{Conclusion}

This paper focuses on the modality missing problem in multimodal punctuation tasks, as unpunctuated sentences are a mixture of audio text and audio-free text in real applications. 
We devise a new unified multimodal punctuation framework, named \method.
\method can learn a hybrid representation for both audio and audio-free corpus, based on which \method allows punctuating both kinds of sentences. We experiment on two real-world datasets, and find that \method surpasses all previous strong multimodal and unimodal baselines by at least 0.8 overall F1 scores. Extensive experiments show that our framework is pervasive and can empower other models to process modality missing sentences. 

\bibliographystyle{IEEEbib}
\bibliography{strings,refs}

\end{document}